\documentclass{article}

\usepackage[numbers,sort&compress]{natbib}

\usepackage[final]{neurips_2023}

\usepackage[utf8]{inputenc} 
\usepackage[T1]{fontenc}    
\usepackage{hyperref}       
\usepackage{url}            
\usepackage{booktabs}       
\usepackage{amsfonts}       
\usepackage{nicefrac}       
\usepackage{microtype}      
\usepackage{xcolor}         

\usepackage{amsmath, amsthm, amssymb}
\usepackage{mathrsfs}
\usepackage{tipa}
\usepackage{graphicx}
\usepackage{multirow} 
\usepackage{wrapfig}
\usepackage{color}

\begin{document}

\title{Structural Similarities Between Language Models and Neural Response Measurements}

\newcommand{\ku}{\spadesuit}
\newcommand{\pu}{\blacklozenge}
\newcommand{\dtu}{\clubsuit}
\newcommand{\cpai}{\text{\normalfont \textipa{C}}}
\newcommand{\pai}{\text{\normalfont \textipa{D}}}

\author{%
  Jiaang Li\thanks{
  Equal Contribution.
  }$^{~,\ku,\cpai}$ 
  \And
  Antonia Karamolegkou$^{\ast,\ku}$ 
  \And
  Yova Kementchedjhieva$^{\ku}$ 
  \And
  Mostafa Abdou$^{\pu}$ 
  \And
  Sune Lehmann$^{\dtu,\pai}$ 
  \And
  Anders Sogaard\thanks{
  Correspondence to \href{mailto:soegaard@di.ku.dk}{soegaard@di.ku.dk}.
  }$^{~,\ku,\pai,\cpai}$ \\
  \And \\
  $^{\ku}$University of Copenhagen \ \
  $^{\pu}$Princeton University \ \
  $^{\dtu}$Technical University of Denmark\\
  $^{\pai}$Pioneer Centre for AI, Denmark\ \
  $^{\cpai}$Center for Philosophy of AI, Denmark
}

\maketitle

\begin{abstract}
Large language models (LLMs) have complicated internal dynamics, but induce representations of words and phrases whose geometry we can study. Human language processing is also opaque, but neural response measurements can provide (noisy) recordings of activation during listening or reading, from which we can extract similar representations of words and phrases. Here we study the extent to which the geometries induced by these representations, share similarities in the context of brain decoding. We find that the larger neural language models get, the more their representations are structurally similar to neural response measurements from brain imaging. Code is available at \url{https://github.com/coastalcph/brainlm}.
\end{abstract}

\section{Introduction}
\label{sec:intro}

Understanding how the brain works has intrigued researchers for many years. This challenge has given rise to the field of brain decoding, where the goal is to interpret the information encoded in the brain while a person is engaged in a specific cognitive task, such as reading or listening to language. By analyzing representations of neural activity across different brain regions, researchers can develop computational models that link specific patterns of brain activity to linguistic elements, such as words or sentences. This direction of research opens avenues for advancing our understanding of neurological disorders, developing innovative treatments, and enhancing the quality of life for individuals with disorders.

\begin{figure}
    \centering
    \includegraphics[width=0.9\textwidth]{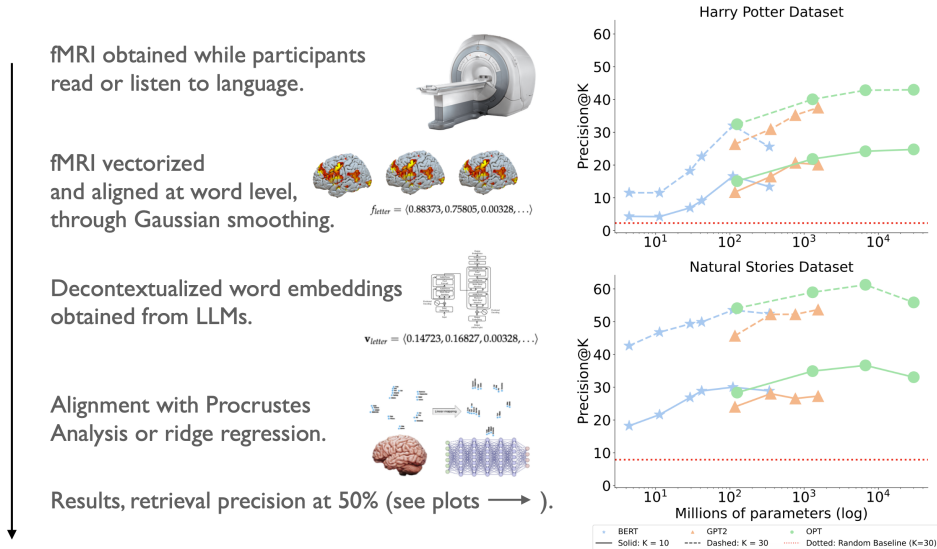}
    \caption{{\bf Experimental flow and main results.} We run experiments with three families of LLMs (comparing LLMs of different sizes within families), two fMRI datasets, and three projection algorithms, and results are the same across all combinations: LLMs converge toward human-like representations, enabling (P@10) retrieval rates of up to 25\%, i.e., a quarter of all concepts can be decoded from the fMRI signals. The datasets, our Gaussian smoothing technique, and the projection methods are described in \S \ref{sec:methods}. Right side: \textbf{Convergence results for three families of LLMs across two datasets, using Procrustes analysis}. Convergence is consistent, and some retrieval rates are remarkably high, decoding almost half of the words correctly to a neighborhood of 30 word forms, which surpasses the random retrieval baselines represented by the dotted \textcolor{red}{red} lines.
    }
    \label{fig:flow}
\end{figure}

In this paper, we investigate the alignment between the representations of words in LLMs and the neural response patterns observed in the human brain during language processing. What emerges is a striking structural similarity between these two sets of representations, manifesting as a geometric congruence in high-dimensional vector spaces. To quantify this alignment, we employ rigorous evaluation methods, including ridge regression, representational similarity analysis (RSA) \citep{Kriegeskorte2008-RSA}, and Procrustes analysis \citep{kementchedjhieva-etal-2018-generalizing} (if $d=d'$). These methodologies enable us to quantify the extent of isomorphism between LLMs and neural responses, e.g., functional magnetic resonance imaging (fMRI). Figure \ref{fig:flow} illustrates the experimental flow and main results.

Artificial intelligence researchers evaluate LLMs by measuring their performance on benchmark data and protocols \citep{51569,cabello2023pokemonchat,doi:10.1073/pnas.2215907120}. Doing so, they aim to infer what LLMs have learned, from how they behave. The methodology is behaviorist and has obvious limitations. We instead suggest exploring the inside of LLMs and our brains -- or, to be precise, their representational~{\em geometries}. Our investigations span various LLM families, word embeddings, and diverse datasets, consistently revealing high degrees of structural congruence. Our main contributions are as follows:
\begin{itemize}
    \item We find a remarkable structural similarity between how words are represented in LLMs, and the neural response measurements of humans reading the same words. The LLM representations of a vocabulary form a geometry in a $d$-dimensional vector space; and the neural response measurements from one or more participants reading these words in a brain scanner, form in a similar way a geometry in a $d'$-dimensional space. 
    \item We present experiments for three families of LLMs (as well as one static embedding method), with two different fMRI datasets, and three evaluation methods (ridge regression, RSA, and Procrustes analysis) to compute the structural similarity (degree of isomorphism) between these two modal geometries. 
    \item Across the board, we see high degrees of isomorphism, enabling decoding or retrieval performance (precision-at-$k$, a.k.a P@$k$) of up to P@10$\approx$25\% (with random performance being P@10$<$1\%).  Word-level brain decoding thus seems feasible particularly as language models increase in size.
\end{itemize}

\section{Related Work}
Over the past decade, researchers have explored the relationship between neural and language representations by predicting text from brain activity \cite{sogaard2016evaluating,ramakrishnan2021non,fereidooni2020understanding}. \citet{pereira2018} were the first to build regression models to predict sentence representations from brain scans. Extending this work, \citet{minnema-herbelot-2019-brain} investigate several metrics to evaluate the decoder performance. Apart from regression, \citet{sun_wang_2019_decoding} also use similarity-based decoders where the decoder is trained to map brain images to distinct sentence representations in both structured and unstructured settings. \citet{affolter2020brain2word} use a neural network model to facilitate the brain-to-word regression decoder, and evaluate on unseen subjects for a more realistic approach. \citet{oota-etal-2022-multi} propose two novel setups using multi-view and cross-view regression decoders that predict semantic concepts and vector representations respectively. \citet{zou-etal-2022-cross} suggest a neural decoder in a cross-modal cloze setting predicting the target word given a contextual prompt. Finally, \citet{Tang2022.09.29.509744} build a decoder that reconstructs continuous language instead of individual words or sentences. Our focus is not on building a better decoding system, but rather on exploring the alignment between neural and language representational spaces within such systems. 
\section{Methodology}
\label{sec:methods}
We begin with a description of our empirical results. We have experimented with three families of LLMs, comparing the word representations they induce to human representations obtained from two different neural response measurement datasets. We use three different comparison methods, leading to a total of 18 experiments, which all confirm the same trend. 
\subsection{Data Description and Pre-processing}
\paragraph{fMRI datasets.}
fMRI is a non-invasive neural response measurement technique that records on a spatial resolution in the region of 1 to 6 millimetres, higher than any other technique. fMRI records activity (blood flow) in the entire network of brain areas engaged when subjects undertake particular tasks. On the downside, fMRI is somewhat susceptible to influences of non-neural changes, and the temporal response is poor relative to the electrical signals that define neuronal communication. To compensate for low temporal resolution, we introduce Gaussian smoothing below. The datasets are: \href{http://www.cs.cmu.edu/~fmri/plosone/}{Harry Potter Dataset} \citep{Wehbe2014-simultaneously} (8 subjects) , and \href{https://osf.io/eq2ba/}{Natural Stories Audio Dataset} \citep{Zhang649939} (19 subjects). Both datasets are publicly available. 

\paragraph{Gaussian smoothing.}
As mentioned, we use Gaussian smoothing to extract word-level neural response measurements in our two datasets. Gaussian smoothing has been used before to study speech-aligned fMRI data \citep{bingel-etal-2016-extracting,Brodoehl2020SurfacebasedAI}. In cases where fMRI data is not collected at the granularity of individual words, we can use Gaussian smoothing to generate word-level fMRI information. For instance, to obtain the fMRI vector for a specific word like "Harry" at a given time point t (Harry$_t$), we can extract the fMRI vectors for a certain timeframe T around t, such as  $t \pm T$  seconds. We then apply Gaussian smoothing to this set of vectors, resulting in a final vector that represents the fMRI information for the word "Harry$_t$".
This approach has potential benefits for fMRI analysis in various applications, such as studies of language processing and cognitive neuroscience. By generating word-level fMRI information using Gaussian smoothing, we can potentially extend the scope from sequence-level to word-level, and improve the interpretability and accuracy of the results obtained from fMRI analyses. Extracting word-level signals differentiates our work from much other work, but was also shown to be crucial in recent work on brain decoding \citep{Tang2022.09.29.509744}.

\subsection{Models}

\paragraph{Auto-regressive models.} 
Auto-regressive models generate output sequences by predicting each element in the sequence based on the previously generated elements. In other words, the output is generated one element at a time, with the model conditioned on the previous output elements. These language models are used to generate text but typically provide slightly worse similarity estimates. We use two auto-regressive language model families: GPT2 \citep{Radford2019LanguageMA} and OPT \citep{zhang2022opt}. 

\begin{table*}
\centering
\caption{The 14 language models used in our experiments. Appendix \ref{sec:implementation} Table~\ref{tab:lm_path} lists the links of LMs.}
\resizebox{0.95\linewidth}{!}{
\begin{tabular}{l|cccrl}
\toprule 
\textbf{LMs} & \textbf{Hidden Layers} & \textbf{Dimension Size} & \textbf{Attention Heads} & \textbf{Total \# of Params} & \textbf{Datasets} \\
\midrule
\multirow{6}{1.1cm}{BERT} & 2 & 128 & 2 & 4.4M & \multirow{6}{8cm}{BooksCorpus \citep{Zhu2015AligningBA}, English Wikipedia \citep{devlin-etal-2019-bert}}\\
 & 4 & 256 & 4 & 11.3M & \\
 & 4 & 512 & 8 & 29.1M & \\
 & 8 & 512 & 8 & 41.7M & \\
 & 12 & 768 & 12 & 110.1M & \\
 & 24 & 1,024 & 16 & 336M & \\
\midrule
\multirow{4}{1.1cm}{GPT2} & 12 & 768 & 12 & 117M & \multirow{4}{8cm}{WebText \citep{Radford2019LanguageMA}} \\
 & 24 & 1,024 & 16 & 345M & \\
 & 36 & 1,280 & 20 & 762M & \\
 & 48 & 1,600 & 25 & 1,542M & \\
\midrule
\multirow{4}{1.1cm}{OPT} & 12 & 768 & 12 & 125M & \multirow{4}{8cm}{BooksCorpus, CC-Stories\citep{Trinh2018ASM}, CCNewsV2\citep{zhang2022opt}, The Pile\citep{DBLP:journals/corr/abs-2101-00027}, Pushshift.io Reddit dataset \citep{DBLP:journals/corr/abs-2001-08435}} \\
 & 24 & 1,024 & 32 & 1.3B & \\
 & 32 & 4,096 & 32 & 6.7B & \\
 & 48 & 7,168 & 56 & 30B & \\
\bottomrule 
\end{tabular}}
\label{tab:LMs_details}
\end{table*}

\paragraph{Non-auto-regressive models.} Non-auto-regressive models are a type of machine learning models that take in an entire input sequence of text and generate a single output vector representation for the entire sequence. Training such models, we mask a fraction of the words in the input text. The language model is then expected to predict the masked words based on the other words in the text. We use the BERT \citep{devlin-etal-2019-bert} family of language models as an example of a non-auto-regressive model family. See more details in Table \ref{tab:LMs_details}.

\paragraph{Language representation.}
In this study, the LLMs we utilized were trained on text segments, so applying these models to individual words in isolation might yield unpredictable results. Instead, our approach involved retrieving naturally occurring instances of these words, along with their surrounding context (i.e., the full sentence), from the fMRI text material. Consequently, we obtain representations from the token positions that align with each word when encoding these sentences, and subsequently average the tokens representations of each word. To ensure decontextualization, we further averaged these representations across different sentences \cite{abdou-etal-2021-language, li2023implications}.

\subsection{Comparison and Projection Methods}
\paragraph{Representational similarity analysis.}
Relational similarity analysis (RSA) is a multivariate analysis technique commonly used in cognitive neuroscience and computational linguistics to compare the similarity between two sets of representations \citep{Kriegeskorte2008-RSA}. RSA can be used to measure the similarity between the neural activity patterns observed in the fMRI data and the representations learned by LLMs. RSA operates by first representing the neural activity and language model features as vectors in a high-dimensional space. The similarity between these vectors is then quantified using a rank-based correlation metric. 
We perform RSA following \citet{lepori-mccoy-2020-picking}. 
Let $X$ and $Y$ be two sets of representations. We calculate their representational dissimilarity matrices (RDMs) as $\vec{D}_X$ and $\vec{D}_Y$, respectively \cite{lepori-mccoy-2020-picking}. We then compare the representational geometries using Spearman's rank correlation coefficient, denoted as $\rho(\vec{D}_X, \vec{D}_Y) $.

\paragraph{Ridge regression.}
Ridge regression is a widely used method in statistics and machine learning to address the issue of multicollinearity, which can arise when there are highly correlated predictor variables in a linear regression model. In contrast to \citet{Toneva2019InterpretingAI}, who utilized ridge regression for encoding fMRI data, our approach focuses on decoding, i.e. predicting language from fMRI. We achieve this by establishing a model that captures the connection between brain signals and individual dimensions within the language model representations.
The models are trained to predict the signal of word $w$ in layer $l$, denoted as $y_{l^w}$, using the vector of fMRI voxels for that word, $x^w$. For each subject and layer $l$, we employ cross-validation to estimate the predictiveness of the fMRI representation of the word in each dimension $i$. In each fold, the fMRI data matrix with total $n$ dimension denoted as $X = x_{w^1}, x_{w^2}, ..., x_{w^n}$, and the semantic vector matrix with $m$ dimension, denoted as $Z = z_{w^1}, z_{w^2}, ..., z_{w^m}$, are split into corresponding training and validation matrices which are individually normalized to have a mean of 0 and a standard deviation of 1 for each dimension across words, ending with training matrices $X^R$ and $Z^{R,l}$, as well as validation matrices $X^V$ and $Z^{V,l}$. Using the training fold, we estimate a model $\theta^{i,l}$ as follows:
\begin{equation}
\arg\min_{\theta^{i,l}} ||z^{R,i} - X^{R}\theta^{i,l}||^2_2 + \lambda^i||\theta^{i,l}||^2_2 \nonumber
\end{equation}
To identify the best $\lambda^i$ for each dimension $i$ that minimizes the nested cross-validation error, we employ a ten-fold nested cross-validation. Subsequently, we estimate $\theta^{i,l}$ using $\lambda^i$ on the entire training fold. Thus, the predictions for each dimension in the validation fold are obtained as $p^l = X^{V}\theta^{i,l}$.

\paragraph{Procrustes analysis.}
We use Procrustes analysis, a form of statistical shape analysis, to align brain fMRI representations with those of language models, using a fMRI-text dictionary (see \S \ref{sec:dictionary}). Procrustes analysis is a method for matching corresponding points in two shapes and finding the transformation (translation, rotation, and scaling) that best aligns them. Under the constraint of orthogonality, we aim to optimize the objective function $\Omega = \min_{R}\Vert RX-Z \Vert _{F}$, subject to $R^{T}R=I$, where matrix $X$ refers to the fMRI matrix and matrix $Z$ refers to representations of words from LLMs. This optimization problem has a closed-form solution given by: $\Omega = UV^{T}$, and $U\Sigma V = \text{SVD}(ZX^T)$, where SVD represents the singular value decomposition. 

\section{Experimental Setup}
\subsection{fMRI-text Dictionary Complementation}
\label{sec:dictionary}
We build a bimodal dictionary that associates fMRI data with corresponding textual information, utilizing fMRI datasets. Considering the context in which words are presented, it becomes evident that the brain's response to a particular word may vary significantly across different sentences. This dynamic response suggests that, within our constructed dictionary, the relationship between fMRI recordings and textual entries exhibits a many-to-one correspondence. We employ a four-fold cross-validation approach that takes into account unique words, thereby preventing any potential train-test leakage. Due to individual differences among subjects, our experiments are conducted based on each subject's responses. We report the averaged results across all subjects.

\subsection{Evaluation - Linear Projection}

To assess the effectiveness of regression and alignment techniques, we employ the P@$k$ metric, which quantifies the ratio of accurate predictions within the top $k$ predictions. This evaluation metric offers a more cautious and robust assessment \cite{karamolegkou2023mapping}. For Procrustes analysis, we induce it from a small set of point pairs and test it on held-out data measuring the P@$k$ \cite{lample2018word}, whereas using the whole point pairs to assess the regression performance. What's more, ensuring congruent dimensionality between the source and target spaces is a crucial prerequisite for successful alignment. In instances where a dimensionality mismatch arises, we employ principal component analysis to reduce the dimensionality of the larger space, ensuring compatibility.

\paragraph{Cross-domain similarity local scaling (CSLS).}
\label{sec:csls}
Nearest neighbor relationships are inherently asymmetric, and high-dimensional spaces can lead to 'hubness,' where some vectors are hubs, while others are anti-hubs \citep{hubsjmach}. To address this issue, \citet{lample2018word} propose a bi-partite neighborhood graph, in which each word of a given dictionary is connected to its $K$ nearest neighbors in the other language. They use cross-domain similarity local scaling (CSLS) to evaluate the similarity between mapped source and target words, which improves upon traditional nearest neighbor methods \cite{lample2018word}. We use CSLS to calculate P@$k$.

\begin{wraptable}{l}{0.6\textwidth}
\vspace{-7mm}
\centering
\caption{\label{tab:precision-baseline} Two different P@$k$ baselines with $k\in\{1,5,10,30,50,100\}$ of two datasets. The random retrieval baselines are calculated by the U.W. in stimulus content, respectively. U.W. = the number of unique words.}
\resizebox{1\linewidth}{!}{
\begin{tabular}{l|r|rrrrrr}
\toprule
Datasets & U.W. & P@1 & P@5 & P@10 & P@30 & P@50 & P@100\\
\midrule 
Harry Potter$_{Random}$ & \multirow{2}{*}{1291} & 0.08\% & 0.39\% & 0.77\%  & 2.32\%  &  3.87\%  &  7.74\%\\
Harry Potter$_{FastText}$ &  & 0.36\% & 3.66\% & 6.43\% & 12.76\% & 17.22\% & 26.52\% \\
\midrule
Natural Stories$_{Random}$ & \multirow{2}{*}{381} & 0.26\% & 1.31\% & 2.62\% & 7.87\% & 13.12\% & 26.25\%  \\
Natural Stories$_{FastText}$ &  & 0.00\% & 1.88\% & 5.62\% & 17.27\% & 24.24\% & 39.89\%  \\
\bottomrule 
\end{tabular}}
\end{wraptable}

\paragraph{Random retrieval baseline.}
P@$k$ is a metric that quantifies the proportion of words for which the LLM's representation serves as one of the k-nearest neighbors to the corresponding fMRI encoding. In essence, word-level decoding involves a straightforward nearest-neighbor retrieval process within the projected space. It's crucial to note that our target vector space, which represents the language model, contains hundreds of vectors. This feature sets our random baseline P@$1<0.1\%$. Our target space of the text material in fMRI datasets makes the random retrieval baseline: $\text{P@1} = \frac{1}{N} \sum_{i=1}^{N} \frac{1}{U} \times 100$\%, where $N$ represents the total number of unique words; $i$ iterates over all words in the material; $U$ refers to the total number of unique words.

\paragraph{FastText baseline.}
The random retrieval baseline serves as a crucial benchmark, enabling us to assess the efficacy of aligning representations between the two modalities in the complete absence of any discernible signal (i.e., by chance). Nevertheless, surpassing the random baseline, strictly speaking, does not definitively establish that contemporary LLMs are inherently more aligned with neural response measurements. To address this concern, we conducted a secondary baseline alignment experiment by aligning fMRI recordings with word representations from fastText \cite{bojanowski2017enriching}. Further details can be found in Table \ref{tab:precision-baseline}. In practical applications, our mappings exhibit significantly higher precision, reflecting the inherent structural similarities between the language model and human brains.

\section{Results \& Discussion} 
\subsection{Main Results.}

\begin{figure}
\centering
\includegraphics[width=0.95\textwidth]{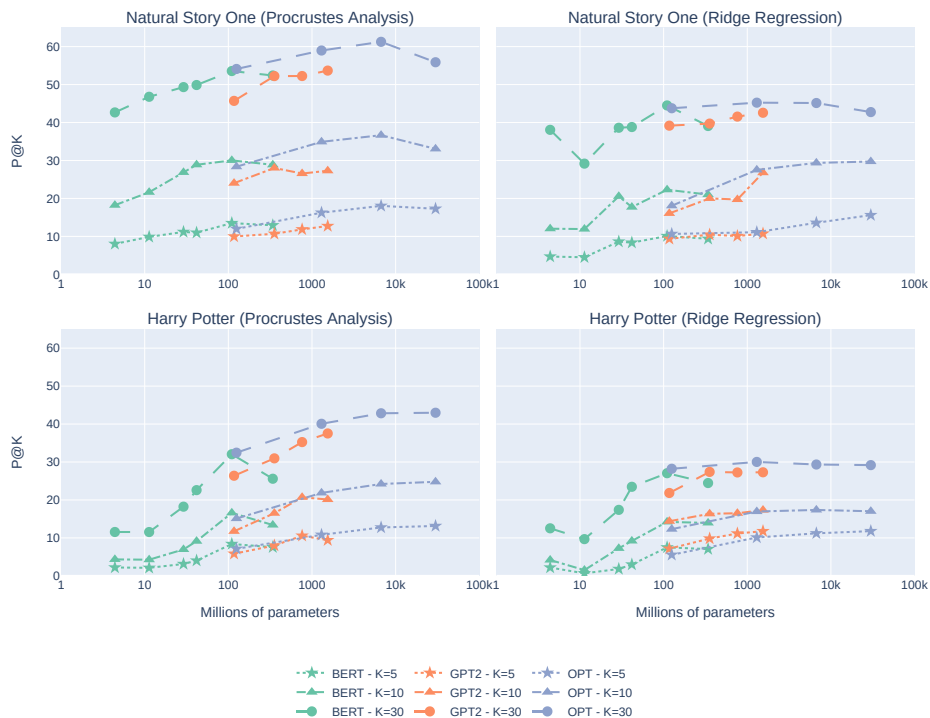}
\caption{\textbf{Convergence Results for Three Families of Language Models on Two Datasets}. 
The task for Harry Potter dataset here is:  Given a neural response, which word in a vocabulary of {1,291} words, was read at the time the response was recorded? Random retrieval baseline P@10 is {\em less} than $1\%$, while the FastText baseline P@10 is {\em less} than $7\%$. See more details of baselines in Table \ref{tab:precision-baseline}.
\label{fig:main}}
\end{figure}

\begin{wrapfigure}{r}{0.6\textwidth}
    \centering
    \vspace{-7mm}
    \includegraphics[width=0.95\linewidth]{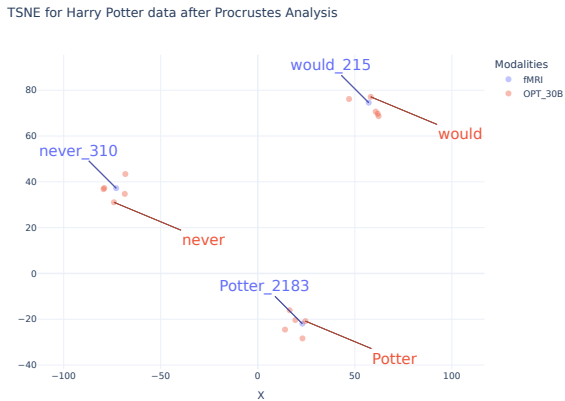}
    \caption{{\bf t-SNE plot of fMRI and LLM representations} using OPT-30B (large, uncased) over select target words from Harry Potter dataset. We evaluate the retrieval performance of our alignments using P@$k$, which measures the ratio of word tokens $w_i$, e.g., for $k=5$, `\texttt{Potter}$_{2183}$', whose fMRI representations are projected into the LLM space such that the LLM representation for the word type $w$, e.g., {\em Potter}, is among the $5$-nearest neighbors of $w_i$. In this case, the neural activity associated with `\texttt{never}$_{313}$' is not read as {\em never} directly -- but still with {\em never} as the top-5 guess. That said, the words {\em Potter} and {\em would} are decoded correctly by our alignment (top-1 guess or P@1).}
    \label{fig:tsne}
    \vspace{-7mm}
\end{wrapfigure}

Our main results are presented in Figure~\ref{fig:main} which illustrates the averaged results across all subjects, and concern the convergence of three families of LLMs on representations that are remarkably similar to those seen in neural response measurements. These results are consistent across two fMRI datasets and three mapping methods. See Appendix \ref{sec:more-results} (Figure \ref{fig:rsa-convergence}) for similar results with RSA.
The scores are plotted by model size, showing the convergence toward brain-like representations as LLMs increase in size. The best scores indicate that LLMs up to {1.5B parameters} can achieve alignments such that a bit more than 1 in 5 words are decoded correctly,\footnote{The reason we count P@5 or P@10 as correct decoding is that a neighborhood of 5-10 words will tend to consist of inflections of the same lemma or synonymous words \citep{kementchedjhieva-etal-2019-lost}. P@1 would amount to guessing the lemma, the exact inflection, and the correct spelling variant.} and a bit more than 2 in 5 almost correctly (within neighborhoods of 20-30 word forms). To gain a qualitative sense of the alignment between brain signal and LLMs representations, see Figure \ref{fig:tsne}. 
The results are obtained with limited supervision for learning the mapping. In fact, we only rely on {950} data points to induce this linear projection, a small number given the high dimensionality of the derived word representations; see \S\ref{sec:methods} for details.

\subsection{Discussion}

Our findings reveal a strong similarity between language model word representations and human brain responses to language stimuli. As these neural language models expand in size, their representations become more akin to the patterns observed in neural responses from the fMRI scans. This discovery points to the development of human-like representations within these large-scale language models, offering valuable insights into the intricate relationship between artificial intelligence and human cognitive processes.

\paragraph{Newman's objection?}

Philosophers argue whether structural similarities (isomorphisms, homomorphisms, etc.) between representations and what is represented, are sufficient for content \citep{Shea2007-SHECAI,mollo2023vector}. Their concerns have their origin in Newman's objection to Russellian structural realism \citep{Newman1928-NEWMRC}. 

Briefly put, Newman showed that structuralist descriptions that abstract away from all but the logical structure, and simply assert the existence of a relation that induces a graph isomorphism between the representation, and what is represented, are indeed trivial. Any LLM will, in other words, induce word representations such that the nearest neighbor graph over the word vocabulary $\mathcal{V}$ such that there exists a relation that is isomorphic to that graph. \citet{mollo2023vector}, for example, bring up Newman's objection and write: 

\begin{quote}
    philosophical work on theories of representational content has
long established [\ldots that m]orphisms
between two sets of objects or properties are trivial to find, and rely solely on the existence of
morphisms between internal representations and structured domains in the world could lead to a trivialisation of the notions of representation and meaning \ldots
\end{quote}

However, Newman's objection only holds if all there is posited is the existence of {\em some} relation. If the relations are properly restricted, isomorphism is far from trivial. One observation that goes all the way back to Carnap's {\it Aufbau} \citep{Carnap1967-CARTLS-3}\footnote{Russell arguably had a similar response \citep{Pashby2015-PASURR}.} is: Structural similarities are generally trivial to obtain, but if the relations (distances in the vector space) serve a purpose (do work for the system), structural similarities can ground content. 

Structural similarity is evidently sufficient to solve semantic problems, such as bilingual dictionary induction \citep{4264a46fd9e846e4a704b2d13002e521} or multi-modal alignment \citep{li2023implications}. The fact that fMRI vectors exhibit structural similarities to LLMs (and by transitivity, across languages and to computer vision models), is suggestive of such similarities playing a role in grounding. 

In our case, we are not simply positing an isomorphic relation in neural responses. We are positing an isomorphism between two very specific relations: the nearest neighbor graph in the LLM representations, and the nearest neighbor graph in the fMRI data. In fact, these two relations are the {\em same}~relation, something which Newman himself proposed as a remedy to his own objection. It should thus be clear that the result presented here is far from trivial.

\paragraph{Where are LLMs most brain-like?}
\begin{figure}
\centering
\includegraphics[width=0.8\textwidth]{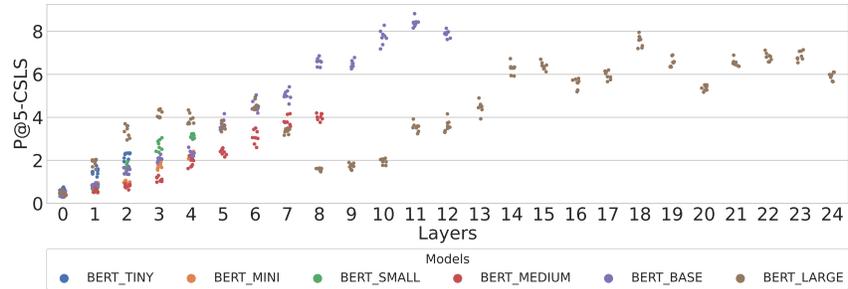}
\caption{\textbf{Alignment precision results across layers with Harry Potter dataset.} The plot shows alignment with fMRI improves with model depth, for BERT and Procrustes Analysis; see Appendix \ref{sec:more-results} for similar plots for other LLMs and projection methods.}
\label{fig:layers}
\end{figure}

We also consider at what layers the different language models align best with the representations extracted from the {fMRI} data. The results presented in Figure~\ref{fig:layers} and the Appendix \ref{sec:more-results} are unambiguous and show that deeper representations align better with neural response measurements. This holds across all architectures and model sizes. 

Interestingly, the alignment improvements at deeper layers do not wear off to reach a plateau. Our results, in fact, suggest that better alignment results can be achieved by training even deeper models. This may also explain the strong correlation between depth and generalization often observed in the literature \citep{goodfellow2014multidigit}.

It has generally been found that the inner-most layers in LLMs encode for syntax, whereas the outer layers encode for semantics and pragmatics. One way to understand our results is therefore that similarities between representations in human brains and LLMs are predominantly driven by semantics and pragmatics. 

\section{Conclusion}
We presented a series of experiments showing that across three families of language models, word representations converge toward being structurally similar to human neural responses. The larger and better the language models get, the more their representations align with human representations. This result holds across datasets and three evaluation methods. We have discussed the philosophical significance of this result, including why Newman's objection does not apply.
\section{Limitations}
Our study demonstrates the precise mapping of neural response measurements to language model representation spaces through supervised learning. However, our findings are subject to certain constraints. The utilization of fMRI signals with limited temporal resolution, albeit partially mitigated through Gaussian smoothing, may introduce potential confounds. Additionally, our primary focus on the English language narrows the generalization ability of our results to languages with different linguistic structures. Furthermore, relying on a single participant for each alignment may introduce individual variability that could influence our conclusions. Moreover, our paper emphasizes the philosophical interpretation of the linear mapping results, leaving the technical aspects of this alignment largely unexplored. To ensure the robustness and broader applicability of our findings, future research should encompass diverse languages, participant groups, and delve deeper into the technical underpinnings of the observed alignment between neural responses and language model representations.
\section{Ethics}
In our research, we analyze two publicly available fMRI datasets (\href{http://www.cs.cmu.edu/~fmri/plosone/}{Harry Potter Dataset} and \href{https://osf.io/eq2ba/}{Natural Stories Dataset}). We did not collect any new dataset for our study. We encourage readers to refer to the terms of use provided by the respective dataset sources for a more comprehensive understanding of their ethical guidelines and data usage policies. We do not foresee any harmful uses of this technology.

\begin{ack}
Thanks to the anonymous reviewers for their helpful feedback. Jiaang Li is supported by Carlsberg Research Foundation (grant CF22-1432). Antonia Karamolegkou is supported by the Onassis Foundation - Scholarship ID: F ZP 017-2/2022-2023’. 
\end{ack}

\bibliography{biblio_scales, custom}
\bibliographystyle{unsrtnat}

\appendix
\section{Implementation} \label{sec:implementation}
Our implementation is based on PyTorch v.1.13.1 \citep{NEURIPS2019_9015} and Transformer v4.25.1 \citep{wolf-etal-2020-transformers} for Python 3.9.13 and builds on code from the repositories in Table~\ref{tab:lm_path}.
\begin{table*}[ht]
\centering
\caption{\label{tab:lm_path} Links of 14 Transformer-based language models used in our experiments.}
\resizebox{0.8\linewidth}{!}{
\begin{tabular}{l|c}
\toprule 
\textbf{LMs} & \textbf{Links}  \\
\midrule 
BERT$_{\textsc{TINY}}$ & \url{https://huggingface.co/google/bert_uncased_L-2_H-128_A-2} \\
BERT$_{\textsc{MINI}}$ & \url{https://huggingface.co/google/bert_uncased_L-4_H-256_A-4} \\ 
BERT$_{\textsc{SMALL}}$ & \url{https://huggingface.co/google/bert_uncased_L-4_H-512_A-8} \\ 
BERT$_{\textsc{MEDIUM}}$ & \url{https://huggingface.co/google/bert_uncased_L-8_H-512_A-8} \\  
BERT$_{\textsc{BASE}}$ & \url{https://huggingface.co/bert-base-uncased} \\
BERT$_{\textsc{LARGE}}$ & \url{https://huggingface.co/bert-large-uncased} \\
\midrule
GPT2$_{\textsc{BASE}}$ & \url{https://huggingface.co/gpt2} \\
GPT2$_{\textsc{MEDIUM}}$ & \url{https://huggingface.co/gpt2-medium}   \\
GPT2$_{\textsc{LARGE}}$ & \url{https://huggingface.co/gpt2-large}   \\
GPT2$_{\textsc{XL}}$ & \url{https://huggingface.co/gpt2-xl} \\
\midrule
OPT$_{\textsc{125M}}$ & \url{https://huggingface.co/facebook/opt-125m}   \\
OPT$_{\textsc{1.3B}}$ & \url{https://huggingface.co/facebook/opt-1.3b} \\
OPT$_{\textsc{6.7B}}$ & \url{https://huggingface.co/facebook/opt-6.7b} \\
OPT$_{\textsc{30B}}$ & \url{https://huggingface.co/facebook/opt-30b} \\
\bottomrule 
\end{tabular}}
\end{table*}

\section{More Results} \label{sec:more-results}
\begin{figure}[ht]
\includegraphics[width=0.9\textwidth]{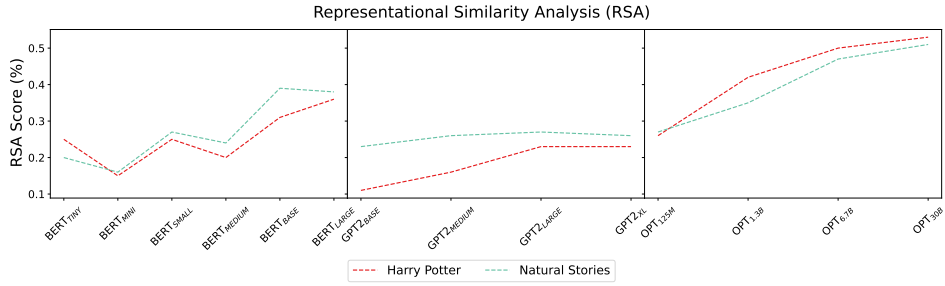}
\caption{\textbf{Convergence results for three families of LLMs using Relational Similarity Analysis}. The correlation score ranges from 0 (no correlation) to 1 (perfect correlation). The plot shows that as the model sizes increase, the representational similarities increase also. \label{fig:rsa-convergence}}
\end{figure}

\begin{figure}[ht]
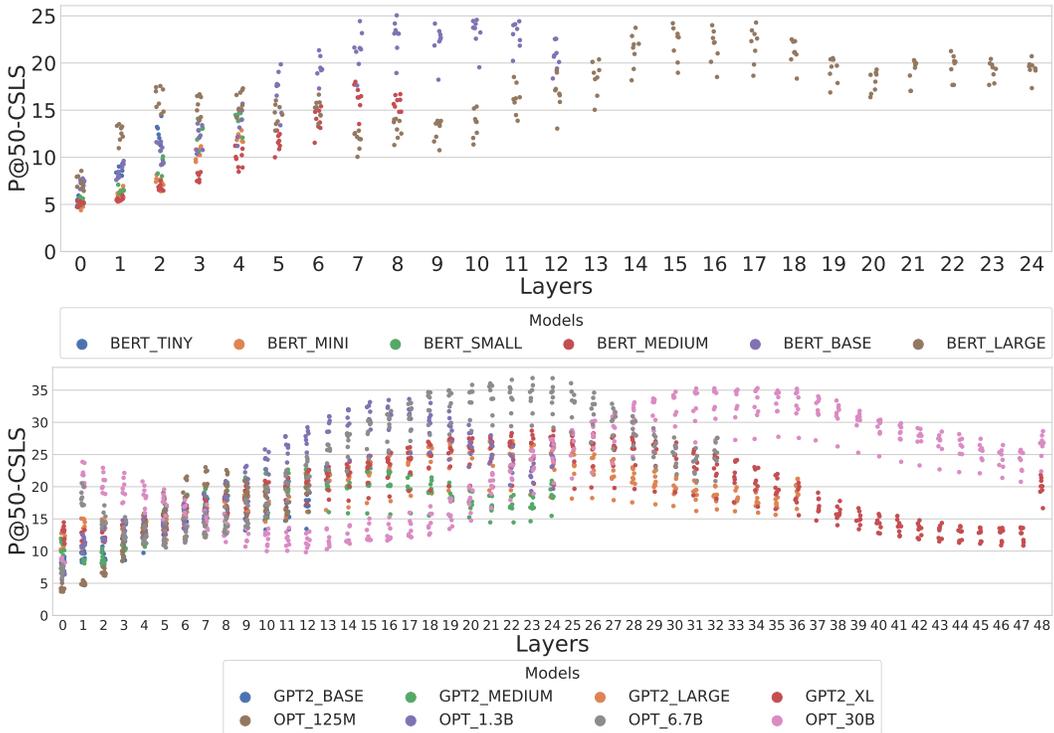

    \centering
    \includegraphics[width=\textwidth]{Figures/HP-token-procrustes-bert-layers-P50-CSLS.pdf}
    \includegraphics[width=\textwidth]{Figures/HP-token-procrustes-gpt-layers-P50-CSLS.pdf}
    \caption{\textbf{Alignment precision results across layers.} The plot shows alignment with fMRI (Harry Potter dataset) improves with model depth for LLMs and Procrustes analysis with Gaussian Random Projection.}
    \label{fig:token-procrustes-layers-rp}
\end{figure}

\end{document}